\documentclass[runningheads]{llncs}
\usepackage{graphicx}
\usepackage{hyperref}
\usepackage{multirow}
\usepackage{booktabs}
\usepackage{amssymb}

\usepackage{multicol}
\usepackage[table]{xcolor}
\usepackage{colortbl}
\usepackage{makecell}
\definecolor{mygray}{gray}{.9}
\usepackage{wrapfig}
\usepackage[misc]{ifsym}
\usepackage{hyperref}

\begin{document}

\title{When 3D Partial Points Meets SAM: Tooth Point Cloud Segmentation with Sparse Labels}
\titlerunning{Tooth point cloud segmentation with extremely sparse labels}
%
%
\authorrunning{Y, Liu et al.}
\author{Yifan Liu \inst{1}, Wuyang Li \inst{1}, Cheng Wang \inst{1}, Hui Chen \inst{2}, Yixuan Yuan\inst{1}\textsuperscript{(\Letter)}
}
\institute{ Department of Electronic Engineering, The Chinese University of Hong Kong, Hong Kong SAR, China \\ \and 
Faculty of Dentistry, The University of Hong Kong, Hong Kong SAR, China 
\\
\email{yxyuan@ee.cuhk.edu.hk} \\
}

\maketitle

\begin{abstract}
Tooth point cloud segmentation is a fundamental task in many orthodontic applications. Current research mainly focuses on fully supervised learning which demands expensive and tedious manual point-wise annotation. Although recent weakly-supervised alternatives are proposed to use weak labels for 3D segmentation and achieve promising results, they tend to fail when the labels are extremely sparse. 
Inspired by the powerful promptable segmentation capability of the Segment Anything Model (SAM), we propose a framework named SAMTooth that leverages such capacity to complement the extremely sparse supervision. To automatically generate appropriate point prompts for SAM, we propose a novel Confidence-aware Prompt Generation strategy, where coarse category predictions are aggregated with confidence-aware filtering. Furthermore, to fully exploit the structural and shape clues in SAM's outputs for assisting the 3D feature learning, we advance a Mask-guided Representation Learning that re-projects the generated tooth masks of SAM into 3D space and constrains these points of different teeth to possess distinguished representations. To demonstrate the effectiveness of the framework, we conduct experiments on the public dataset and surprisingly find with only 0.1\% annotations (one point per tooth), our method can surpass recent weakly supervised methods by a large margin, and the performance is even comparable to the recent fully-supervised methods, showcasing the significant potential of applying SAM to 3D perception tasks with sparse labels. Code is available at \href{https://github.com/CUHK-AIM-Group/SAMTooth}{https://github.com/CUHK-AIM-Group/SAMTooth}.

\end{abstract}
\section{Introduction}

Accurately segmenting teeth in 3D tooth point clouds extracted from Intra-Oral Scanners (IOS) mesh data plays a pivotal role in many orthodontic applications, including detailed analysis of tooth morphology, treatment planning, personalized appliance design, etc \cite{im2022accuracy,liu2022hierarchical,li2022scan,li2022sigma,chen2023medical,wuyang2021joint}. However, existing tooth point cloud segmentation models \cite{zanjani2019mask,xu20183d,lian2020deep,cui2021tsegnet,hao2022toward,xiong2023tsegformer} rely heavily on large annotated datasets for
training, which poses challenges due to the labor-intensive nature of tooth point cloud labeling. For example, it takes around 15 to 30 minutes for an experienced dentist to annotate a half jaw manually \cite{liu2022hierarchical}. This time-consuming process presents a significant obstacle to establishing large-scale datasets with high-quality annotations~\cite{xu2022afsc,he2023h,li2024gtp,chen2022dynamic,li2024gaussianstego} and hinders the generalizability of the diagnosis system~\cite{li2022hierarchical,xu2024immunotherapy,yang2023transferability}.

\begin{figure}[t]
 \centering
  \includegraphics[width=1.0\linewidth]{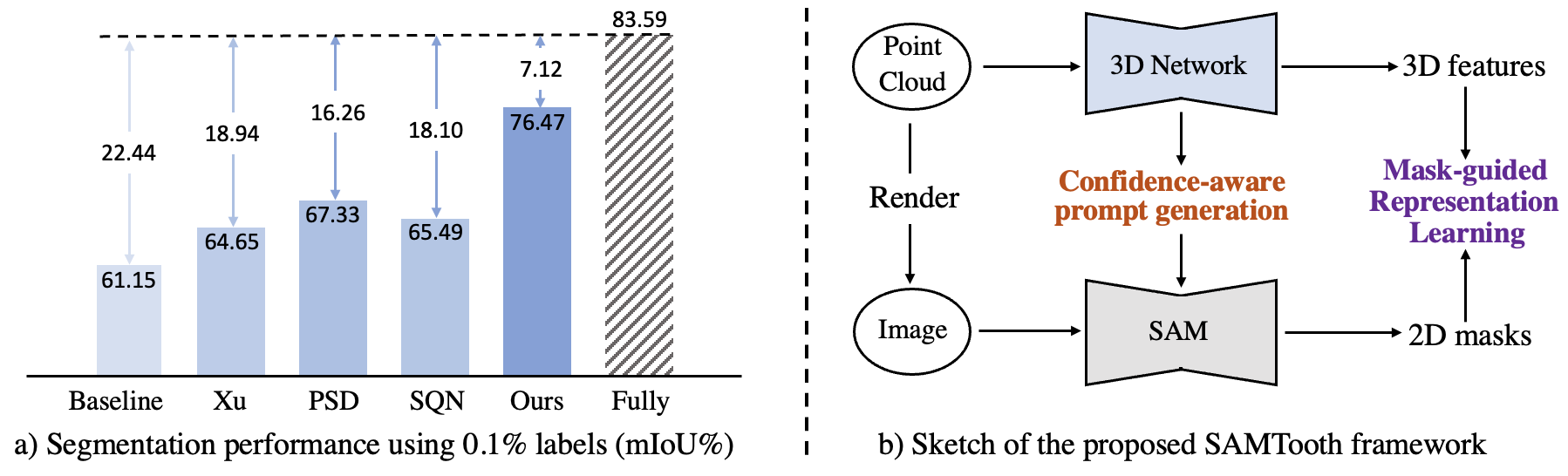}
 \caption{Illustration of a) segmentation performance of various weakly-supervised methods using sparse labels (0.1\%) and b) the proposed SAMTooth framework sketch.}
 \label{fig:introduction}
\end{figure}

To address this issue, there has been a growing interest in investigating weakly-supervised alternatives. Among different types of weak labels (scribbles, boxes, partial points, etc.), partial points stand out as a prospective direction due to the annotating efficiency--it only involves labeling a single or several points for each tooth. Existing partial-points-based methods excavate various training constraints from limited labels, such as perturbation consistency \cite{tarvainen2017mean,zhang2021perturbed}, supervision propagation \cite{xu2020weakly,hu2022sqn}, self-supervised pre-training \cite{li2022hybridcr,jiang2021guided,yang2022mil}, pseudo-labeling \cite{liu2021one,tang2023all,liang2021unsupervised,li2023novel}, etc, which has achieved great progress in reducing the annotation labor. However, as shown in Fig. \ref{fig:introduction}, after increasing the label sparsity to 0.1\% (one point per tooth), we observe that the best entry of existing works \cite{hu2022sqn} only gives marginal 6.18\% performance gains over baseline and yields a 22.44\% mIoU gap compared with fully supervised oracle, indicating that existing works can not perform well \emph{when the labels are extremely sparse}. 

As the first attempt to tackle this issue, we aim to leverage the recent advance of the Segment Anything Model (SAM) \cite{kirillov2023segment}. Trained on a dataset of 11 million images, SAM can generate fine-grained masks given manually defined visual prompts. As shown in Fig. \ref{fig:introduction} a), if we render images from the input 3D model, and feed these images to SAM with adequate prompts, we can get 2D object masks of each tooth. As these masks contain explicit shape information, we can use them to complement the extremely sparse supervision. Nevertheless, it is non-trivial to employ the 2D SAM to assist the 3D task directly, which is attributed to the two issues. Firstly, it is tough to prompt 2D SAM automatically to generate the desired masks. The quality of SAM masks heavily relies on appropriate prompts provided by humans, while incorporating human input during model training is not feasible. Secondly, given the significant disparity between 2D images and 3D point clouds, it is challenging to effectively utilize the 2D masks generated by SAM to enhance model learning in the 3D domain~\cite{liu2024endogaussian,li2024endora,liu2024stereo,liu2024lgs,li2024endosparse}.

To tackle these two challenges, we propose a novel framework SAMTooth, for tooth point cloud segmentation with extremely sparse labels. As shown in Fig. \ref{fig:introduction} b), the framework consists of two paradigms, including \textit{Confidence-aware Prompt Generation} (CPG) and \textit{Mask-guided Representation Learning} (MRL). To automatically generate appropriate prompts for SAM to use, we propose CPG to aggregate the points of each predicted tooth and project the results to the image plane. As the point predictions may be noisy, the point-wise confidence is further estimated to filter unreliable aggregating candidates. To fully leverage the outputs of SAM for 3D feature learning, we advance MRL to re-project the pixels of SAM's outputs into the 3D space and leverage the contrastive learning to provide training constraints. Considering the background points should also be constrained, we also compute a background mask from SAM's object masks and impose explicit supervision. Extensive experiments demonstrate that SAMTooth can outperform other weakly supervised methods by a large margin and is even comparable to recent fully-supervised methods using only 0.1\% annotations.

\section{Method}

\begin{figure}[t]
 \centering
  \includegraphics[width=1.0\linewidth]{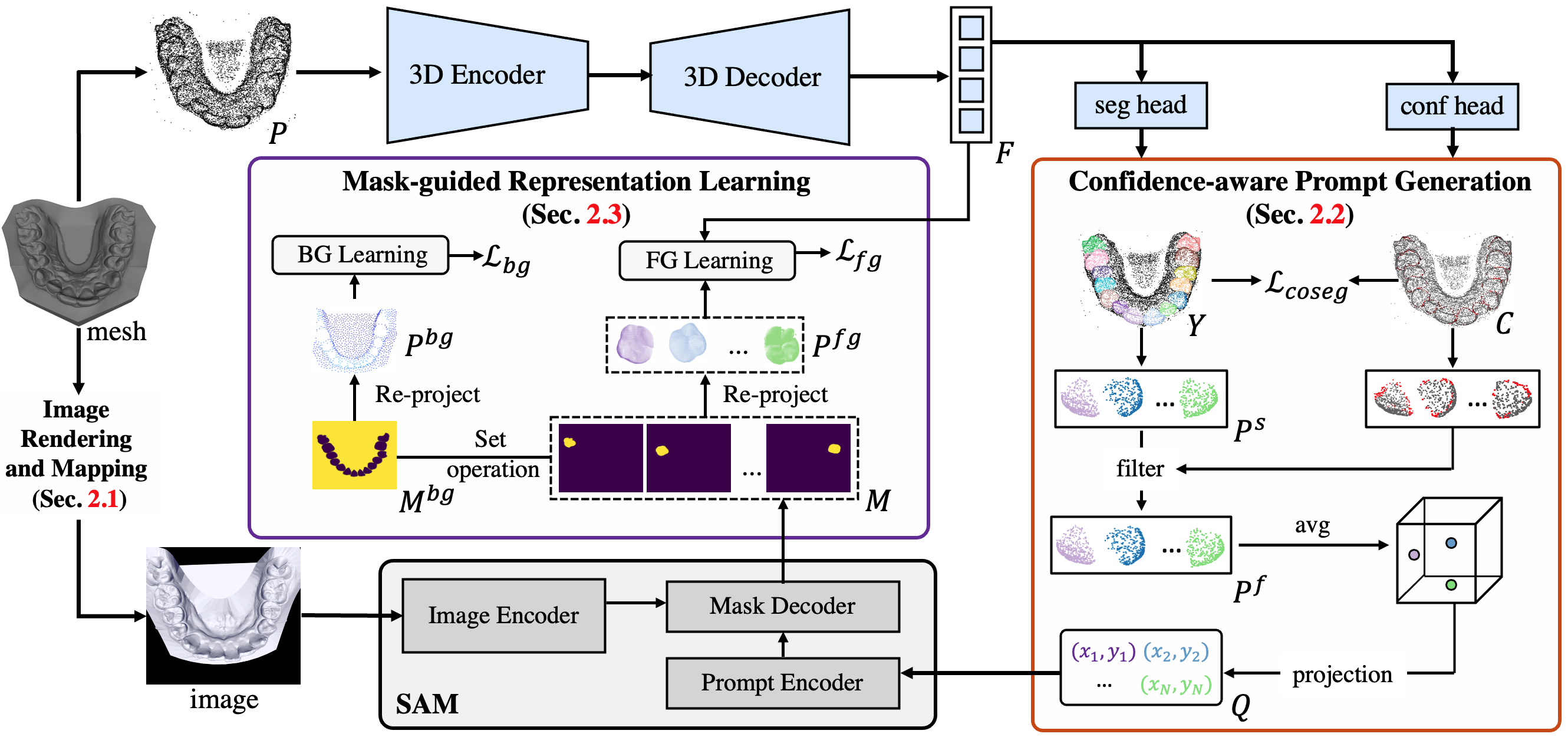}
 \caption{Illustration of the proposed SAMTooth framework.}
 \label{fig:framework}
\end{figure}

Our framework is designed for weakly-supervised tooth point cloud segmentation, by leveraging the zero-shot capacity of visual foundation model SAM. As shown in Fig. \ref{fig:framework}, it begins with Image Rendering and Mapping (Sec. \ref{rendering&mapping}) to render images from the input scan and build the mapping between 3D points and 2D pixels. Then, the input point cloud $P$ is passed to the 3D segmentation network to get coarse predictions $Y$ and point-wise confidence $C$, which are further passed to Confidence-aware Prompt Generation (Sec. \ref{method:CPG}) to generate adequate point prompts for SAM. After that, SAM processes the generated prompts and rendered images to get object masks $M$, which are used to constrain the 3D features by Mask-guided Representation Learning (Sec. \ref{method:MRL}). The whole framework is optimized by the segmentation constraints and complementary constraints from SAM's outputs (Sec. \ref{method:optim}).

\subsection{Image Rendering and Mapping}
\label{rendering&mapping}
To leverage SAM's outputs for 3D representation learning, we first render images from the 3D IOS mesh as SAM's input. We choose to render from the mesh rather than the point cloud as the mesh contains more textural details and is always available in orthodontic applications. Based on the imaging principle of the pinhole camera, the projected coordinates of each point can be obtained by:
\begin{equation}
\label{Eq:2d-3d}
    [u, v, 1]^{T}=1/z\cdot K\cdot T \cdot[x, y, z, 1]^{T},
\end{equation}
where $[u,v]^{T}$ and $[x,y,z]^{T}$ are the 2D and 3D coordinates. $K$ and $T$ are the manually defined camera's intrinsic and extrinsic matrices. By using Eq. \ref{Eq:2d-3d}, 2D images can be rendered from the 3D IOS mesh and 2D pixels can also be projected from the 3D space.

\subsection{Confidence-aware Prompt Generation}
\label{method:CPG}
SAM relies on adequate prompts to generate high-quality object masks, which would further influence the subsequent 3D representation learning. Therefore, a suitable prompt-generation strategy should be carefully designed. In this regard, we propose an automatic prompt generation strategy that gets prompts by aggregating 3D points of each coarsely predicted tooth, accompanied by a confidence-aware filtering step to discard those unconfident tooth predictions that would lead to ambiguous prompts.

\noindent \textbf{Point-wise Confidence Estimation.} In addition to the original segmentation head, we add a confidence head consisting of several MLP and BN layers. Its role is to estimate point-wise confidence values $C=\{c_1, ..., c_N\}\in \mathbb{R}^{N}$. To train the two heads simultaneously, we constrain their outputs using the confidence-aware segmentation loss $\mathcal{L}_{coseg}$ on the labeled point set $P_{label}$:
\begin{equation}
\label{Eq:seg-loss}
    \mathcal{L}_{coseg}=\frac{1}{N}\sum_{p_{i}\in P_{label}}c_{i} \cdot \mathcal{L}_{CE}(y_{i}, y^{gt}_{i}) + (1-c_{i})^{2},
\end{equation}
where $\mathcal{L}_{CE}$ represents the cross-entropy loss between predictions $y_{i}\in Y$ and ground truth $y^{gt}_{i}\in Y^{gt}$. For a certain point $p_{i}$ that the network is confident about the prediction, the estimated $c_{i}$ can be large to reduce the second term in Eq. \ref{Eq:seg-loss} since the first term is already small enough, and vice versa. Therefore, $L_{coseg}$ can encourage the model to generate large $c_{i}$ for confident predictions and small $c_{i}$ for unconfident ones.

\noindent \textbf{Confidence-aware Prompt Generation.} To generate suitable point prompts for each tooth, we first divide the point cloud $P$ into $G$ subgroups $\{P^{s1}, ...,P^{sG}\}$, where each subgroup shares the same category predictions.
Considering noisy points exist in the coarse predictions, we thus use the estimated point-wise confidence as a metric and filter those noisy predictions in a subgroup using a threshold $\tau$, obtaining filtered subgroups $\{P^{f1}, ..., P^{fG}\}$, which $P^{fi}=\{p_{j}|c_{j}>\tau, p_{j}\in P^{si}\}$.
It is easily observed that each subgroup coarsely represents the point cloud of a certain tooth, thus we can get prompts $Q=\{q_1, ...,q_G\}$ by averaging the points belonging to the same subgroup and projecting them to the image coordinates:
\begin{equation}
    q_{i}=Proj(\frac{1}{|P^{fi}|}\sum_{p_{k}\in P^{fi}}p_{k}),
\end{equation}
where $Proj(\cdot)$ projects the 3D coordinates into the 2D image plane as described in Eq. \ref{Eq:2d-3d}. With the guidance of the estimated point-wise confidence, the generated prompts are observed to be closer to the tooth center compared to the simple aggregation, which is more adequate for SAM to produce accurate object masks. 

\subsection{Mask-guided Representation Learning}
\label{method:MRL}
With appropriate point prompts, SAM can generate precise object masks, from which we expect to excavate more constraints to complement the sparse supervision of the 3D model. To this end, we propose to re-project 2D object masks into 3D space and utilize contrastive learning for foreground feature discrimination. Considering the background points, i.e., gingiva should also be constrained, we further compute a background mask from the foreground ones and regularize the corresponding features.

\noindent\textbf{Foreground Learning.}
With previously projected 2D prompts $Q=\{q_{1},...,q_{G}\}$, where $G$ denotes the number of prompts, we can get SAM's output object masks $M=\{m_{1},...,m_{G}\}$, where $m_{i}\in\{0,1\}^{H\times W}$ is the binary mask of a certain tooth. Then, we extract coordinates of object pixels in $m_{i}$ as $coord_{i}=\{(h_{j}, w_{j})|m_{i}(h_{j}, w_{j})=1\}$.
After that, pixels in $coord_{i}$ are re-projected to the 3D space, generating the re-projected 3D subgroup $P^{fgi}$ for each $m_{i}$:
\begin{equation}
    P^{fgi}=\{p_{i}^{3d}|p_{i}^{3d}=ReProj(p_{i}^{2d}), p_{i}^{2d}\in coord_{i}\},
\end{equation}
Doing so for each mask $m_{i}$, we can get the re-projected subgroup set $P^{fg}=\{P^{fg1},...,P^{fgG}\}$. As subgroups should contain points of different categories, i.e., different teeth, we leverage contrastive learning to encourage the 3D features among different subgroups distinguishable. Specifically, we extract 3D features of each point in each group, passing them to two consecutive MLP layers with BN, composing $F^{fg}=\{F^{fg1},...,F^{fgG}\}$. Then, contrastive loss $\mathcal{L}_{fg}$ is imposed on these features:
\begin{equation}
\label{Eq:contra}
    \mathcal{L}_{fg}=-C\sum_{i}\sum_{j}log\frac{exp(f_{i}^{T}f_{j}/t)}{\sum_{k}exp(f_{i}^{T}f_{k}/t)},
\end{equation}
where $f_{i}, f_{j}$ are in the same subgroup,  $f_{i}, f_{k}$ are in the different one, $C$ is the normalization constant, and $t$ is the temperature. The role of Eq. \ref{Eq:contra} can be treated as the complementary supervision to the extremely sparse labels, which provides massive constraints on the unlabeled points.

\noindent \textbf{Background Learning.}
To further constrain the background, i.e., gingiva features, we compute the background mask $M^{bg}$ by eliminating the combination of the pixels in foreground masks $M=\{m_{1},...,m_{G}\}$ generated by SAM:
\begin{equation}
    M^{bg}= \overline{m_{1}}\odot \overline{m_{2}} \odot ... \odot \overline{m_{G}}.
\end{equation}
Then similarly to foreground masks, we also re-project coordinates of pixels in $M^{bg}$ into 3D space to get a background group $P^{bg}$. As we already know these points should belong to the background class, we can directly constrain the predictions of these background features $F^{bg}$ as 0 (the background label):
\begin{equation}
\mathcal{L}_{bg}=CrossEntropy(SegHead(F^{bg}), 0).
\end{equation}

\subsection{Model Optimization}
\label{method:optim}
During training, we first warm up the network using the confidence-ware segmentation loss $\mathcal{L}_{coseg}$ for $T$ epochs, enabling the network to generate coarse segmentation results and point-wise confidence. Then, guided by the output masks of SAM, $\mathcal{L}_{fg}$ and $\mathcal{L}_{bg}$ are used to constrain the foreground and the background 3D features, respectively. The overall optimization objective is:
\begin{equation}
\label{Eq:loss}
    \mathcal{L}=\lambda_{1}\mathcal{L}_{coseg} + (\lambda_{2}\mathcal{L}_{fg} + \lambda_{3}\mathcal{L}_{bg})\cdot [t>T],
\end{equation}
where $t$ is the current epoch and $[\cdot]$ is an indicator function that equals 1 if the statement is true else 0.

\section{Experiments}
\subsection{Experiment settings}

\noindent\textbf{Datasets and evaluation}
To evaluate the effectiveness of our proposed method, we conduct experiments on the public 3DTeethSeg \cite{ben20233dteethseg} dataset. The tooth identification follows the FDI World Dental Federation notation. The 3DTeethSeg \cite{ben20233dteethseg} is a publicly available tooth segmentation dataset, which contains 1,800 available 3D IOS scans obtained from 900 patients, following a real-world patient age distribution. To make a fair comparison, we use the same split in all experiments where 1,080 scans are randomly selected for training, 360 ones for validation, and the remaining ones for testing. Following previous tooth segmentation methods \cite{liu2023grab,liu2022edge,cui2021tsegnet}, we use the Jaccard Index (also known as mIoU), the Dice Similarity Coefficient (DSC), and the point-wise classification accuracy (Acc).

\noindent \textbf{Implementation details}
We adopt standard ViT-B/16 in \cite{yu2022point} as the segmentation backbone. Our framework is trained with AdamW optimizer with a 5e-4 learning rate, 8 batch size, and weight decay of 0.05. We empirically set the confidence thresh-hold $\tau$ as 0.6, temperature $t$ as 0.1, warmup epoch $T$ as 10, and the loss weight $\lambda_{1/2/3}$ in \ref{Eq:loss} as 1/0.1/0.01, respectively. Following previous works \cite{liu2022edge,liu2023grab}, we sample 16,000 points from the IOS scan to compose the input point cloud and use the three-neighbor-interpolation strategy to upsample the predictions to the original size during the evaluation \cite{cui2021tsegnet}.

\subsection{Main results}
\begin{figure}[tbp]
    \centering 
    \includegraphics[width=12cm]{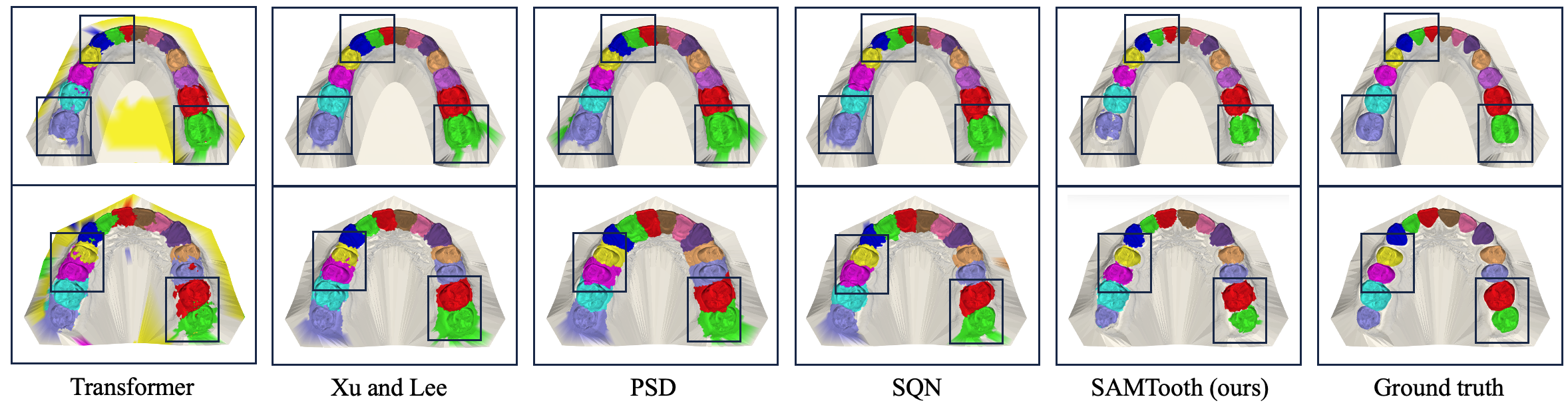}
    \caption{Results comparison on 3DTeethSeg among previous methods and ours.}
    \label{fig:comparison}
\end{figure}

To make a fair comparison with recent state-of-the-art works \cite{laine2016temporal,tarvainen2017mean,xu2020weakly,zhang2021perturbed,hu2022sqn}, we use the same backbone and we re-produce their methods based on the official repositories.  We present the comparison results in \ref{exps:3DTeethSeg-results}. SAMTooth achieves 76.47\% mIoU and 86.64\% mAcc, outperforming previous methods by a large margin. In particular, SAMTooth surpasses II-Model \cite{laine2016temporal}, MT \cite{tarvainen2017mean}, Xu and Lee \cite{xu2020weakly}, PSD \cite{zhang2021perturbed}, and SQN \cite{hu2022sqn} by 15.32\%, 12.47\%, 6.47\%, 11.35\%, 11.82\%, 9.14\%, and 10.98\% in mIoU, respectively. It is also worth noting that with only 0.1\% annotations, SAMTooth can achieve comparable performance (76.47\% vs 83.59\% mIoU) with the fully supervised baseline, which reveals the effectiveness of the proposed framework, and also shows the great potential of SAM for providing training signals for tooth point cloud segmentation with limited labels. We also provide qualitative comparisons in Fig. \ref{fig:comparison}. It can be observed that our method can deliver better segmentation results around boundary regions (black boxes), compared to methods training using other weakly-supervised methods.

\begin{table*}[tbp]
\centering
\scriptsize
\caption{Quantitative results of different methods on the 3DTeethSeg
dataset. The best and second best results are \textbf{bold} and \underline{underlined}.}
\label{exps:3DTeethSeg-results}
\tabcolsep=0.8mm
\renewcommand\arraystretch{1.0}
\begin{tabular}{c|c|ccccc|>{\columncolor{gray!10}}c|>{\columncolor{gray!10}}c}
\hline
Ratio & Methods & Incisor & Canine & Premolar & Molar & Gingiva & mIoU\% & mAcc\% \\
\hline
\multirow{3}{*}{100\%} & PointNet++ \cite{qi2017pointnet++} & 71.81 & 72.31 & 80.07 & 80.67 & 81.34 & 77.15 & 89.06 \\
& DGCNN \cite{wang2019dynamic} & 78.18 & 78.26 & 80.07 & 78.25 & 75.91 & 78.88 & 87.45 \\
& Transformer \cite{yu2022point} & 83.83 & 84.35 & 83.95 & 81.93 & 89.31 & 83.59 & 91.85 \\
\hline
\multirow{7}{*}{0.1\%} & Transformer \cite{yu2022point} & 58.85 & 60.91 & 62.35 & 62.84 & 61.6 & 61.15 & 74.91 \\
& II-Model \cite{laine2016temporal} 
& 64.64 & 65.24 & 66.52 & 64.39 & 63.34 & 64.70 & 77.30 \\
& MT \cite{tarvainen2017mean}
& 64.29 & 64.42 & 67.14 & 65.13 & 61.97 & 65.12 & 77.00 \\
& Xu and Lee \cite{xu2020weakly}
& 63.51 & 64.53 & 67.09 & 64.41 & 61.52 & 64.65 & 76.63 \\
& PSD \cite{zhang2021perturbed} 
& \underline{67.16} & \underline{67.31} & \underline{69.67} & 66.17 & \underline{65.17} & \underline{67.33} & \underline{78.79} \\
& SQN \cite{hu2022sqn}
& 63.18 & 64.68 & 67.72 & \underline{67.52} & 64.38 & 65.49 & 78.63 \\
& Ours 
& \textbf{75.94} & \textbf{77.33} & \textbf{78.02} & \textbf{73.54} & \textbf{78.52} & \textbf{76.47} & \textbf{86.64} \\
\hline
\end{tabular}
\end{table*}

\subsection{More analysis}
\begin{figure}[tbp]
    \centering 
    \includegraphics[width=12cm]{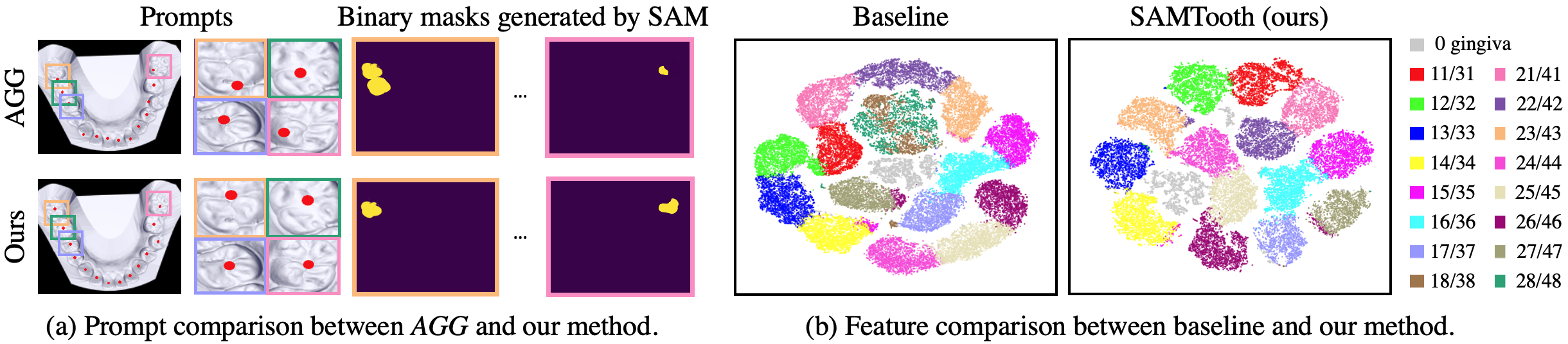}
    \caption{Illustration of (a) prompt comparison and (b) feature comparison.}
    \label{fig:ablation}
\end{figure}

\noindent \textbf{Confidence-aware Prompt Generation.} To evaluate the effectiveness of CPG, we experiment with another prompt generation strategy \textit{AGG}, which obtains point prompts by simple aggregation among each subgroup. As shown in Table. \ref{exps:ablation}, such a simple aggregation strategy would cause a performance drop of 4.63\% in mIoU, revealing the necessity of confidence guidance. We also report qualitative results in Fig. \ref{fig:ablation} (a), from which we observe prompts generated by \textit{AGG} tend to bias~\cite{li2023adjustment} from the center of the tooth, and such prompts would result in mistaken object masks. In contrast, prompts generated by \textit{CPG} are often located around tooth centers and the resulting masks can seamlessly cover each tooth, which can benefit the subsequent representation learning.

\noindent \textbf{Mask-guided Representation Learning.}
Apart from MRL, we experiment with other constraining strategies, including \textit{FL} and \textit{BL} that solely use foreground and background learning. As shown in Table. \ref{exps:ablation}, using \textit{FL} can already outperform the baseline with 12.40\% mIoU gains, due to the complementary constraints for the foreground feature learning. Meanwhile, using \textit{BL} can also bring 3.68\% mIoU advancement. Furthermore, combining \textit{FL} and \textit{BL}, i.e., MRL, can improve the performance with the largest 15.32\% mIoU improvements over the baseline, revealing the effectiveness of MRL. In addition, We present the T-SNE feature visualizations in Fig. \ref{fig:ablation} (b). In general, the features of the baseline are scattered with category mixing, e.g., 11/31, 12/32, and 13/33. In contrast, the features of SAMTooth are more intra-class compact with clear boundaries.

\begin{table}[tbp]
\scriptsize 
\centering
\caption{Ablation studies of CPG and MRL on 3DTeethSeg.}
\label{exps:ablation}
\tabcolsep=1.5mm
\renewcommand\arraystretch{1.0}
\begin{tabular}{c|c|ccccc|>{\columncolor{gray!10}}c}
\hline
Methods & w/ & Incisor & Canine & Premolar & Molar & Gingiva & mIoU \\
\hline
Baseline & - & 57.62 & 61.11 & 63.94 & 63.20 & 72.58 & 61.64 \\
\hline
\multirowcell{2}{Prompt \\ generation} & AGG & 67.83 & 71.09 & 76.48 & 75.51 & 87.94 & 73.11 \\
& CPG & 74.03 & 77.97 & 80.07 & 78.87 & 90.65 & 77.74 \\
\hline
\multirowcell{3}{Mask \\constraints}  & FL & 68.44 & 72.16 & 75.46 & 75.99 & 88.41 & 73.55 \\
& BL & 61.38 & 65.63 & 67.65 & 67.51 & 70.96 & 65.32 \\
& MRL & 74.03 & 77.97 & 80.07 & 78.87 & 90.65 & 77.74 \\
\hline
\end{tabular}
\end{table}

\section{Conclusion}
In this paper, we propose a novel framework for weakly-supervised tooth point cloud segmentation, coined SAMTooth. It leverages the recent advanced promptable foundation model, i.e., SAM, to complement the extremely sparse supervision (one point per tooth).
It adopts a Confidence-aware Prompt Generation
(CPG) to automatically generate precise prompts for SAM to use, guided by the estimated point-level confidence. Then, it leverages Mask-guided Representation Learning (MRL) to achieve maximal utilization of the fine-grained masks generated by SAM. Extensive experiments on two benchmarks show that the proposed method shows significant superiority over existing approaches, showcasing the potential of applying SAM for 3D perception tasks.

%
%
%
\bibliographystyle{splncs04}
\bibliography{paper}




\end{document}